\documentclass[11pt,a4paper]{article}
\pdfoutput=1
\usepackage[hyperref]{naaclhlt2019}
\usepackage{times}
\usepackage{latexsym}

\usepackage{url}

\usepackage{arydshln}
\usepackage{tabularx}
\usepackage{booktabs}
\usepackage{multirow}
\usepackage{amssymb}
\usepackage{amsmath}
\usepackage{pgfplots}
\usepackage{tikz}
\usepackage{textcomp}

\usepackage{todonotes}

\definecolor{c-w2v}{cmyk}{1,0.3968,0,0.2588} 
\definecolor{c-both}{cmyk}{0,0.6175,0.8848,0.1490} 
\definecolor{c-alc}{cmyk}{0.1127,0.6690,0,0.4431} 
\definecolor{c-intra}{cmyk}{0.6765,0.2017,0,0.0667} 
\definecolor{c-inter}{cmyk}{0.3081,0,0.7209,0.3255} 
\definecolor{c-avg}{cmyk}{0,0.8765,0.7099,0.3647}
\definecolor{confred}{rgb}{1,0.48,0.48}
\definecolor{confgreen}{rgb}{0.6,0.9,0.6}
\definecolor{gray}{rgb}{0.5, 0.5, 0.5}

\makeatletter
\def\adl@drawiv#1#2#3{%
        \hskip.5\tabcolsep
        \xleaders#3{#2.5\@tempdimb #1{1}#2.5\@tempdimb}%
                #2\z@ plus1fil minus1fil\relax
        \hskip.5\tabcolsep}
\newcommand{\cdashlinelr}[1]{%
  \noalign{\vskip\aboverulesep
           \global\let\@dashdrawstore\adl@draw
           \global\let\adl@draw\adl@drawiv}
  \cdashline{#1}
  \noalign{\global\let\adl@draw\@dashdrawstore
           \vskip\belowrulesep}}
\makeatother

\aclfinalcopy 
\def\aclpaperid{747} 

\title{Attentive Mimicking:\\
Better Word Embeddings by Attending to
Informative Contexts}

\author{Timo Schick \\
  Sulzer GmbH \\
  Munich, Germany \\
  {\tt timo.schick@sulzer.de} \\\And
  Hinrich Sch\"utze \\
  Center for Information and Language Processing \\
  LMU Munich, Germany \\
  {\tt inquiries@cislmu.org} \\}

\date{}

\newcounter{notecounter}

\newcommand{\enoteson}{\long\gdef\enote##1##2{{
\stepcounter{notecounter}
{\large\bf
\hspace{1cm}\arabic{notecounter} $<<<$ ##1: ##2
$>>>$\hspace{1cm}}}}}
\enoteson

\def\secref#1{\S\ref{sec:#1}}
\def\seclabel#1{\label{sec:#1}}
\def\eqref#1{Eq.~\ref{eqn:#1}}

\begin{document}
\nocite{kingma2014adam}
\maketitle
\begin{abstract}
Learning high-quality embeddings for rare words is a hard
problem because of sparse context information. Mimicking
\citep{pinter2017mimicking} has been proposed as a solution:
given embeddings learned by a standard algorithm, a model is first trained to reproduce embeddings of frequent words from their surface form and then used to compute embeddings
for rare words. In this paper, we introduce \emph{attentive
mimicking}: the mimicking model is given access not only to a word's surface form, but also to all
available contexts and learns to attend to the most
informative and reliable contexts for computing an embedding. In an
evaluation on four tasks, we show that attentive mimicking
outperforms previous work for both rare and medium-frequency
words. Thus, compared to previous work, attentive mimicking improves
embeddings for a much larger part of the vocabulary, including the medium-frequency range. 
\end{abstract}

\section{Introduction}

Word embeddings have
led to large performance gains in
natural language processing (NLP). However,
embedding methods generally need
many observations of a word to
learn a good representation for it.

One way to overcome this limitation and improve embeddings
of infrequent words is to incorporate surface-form
information into learning. This can either be
done directly
\cite{wieting2016charagram,bojanowski2016enriching,salle2018incorporating},
or a two-step process is employed:
first, an embedding model
is trained on the word level and then,
surface-form information is used either to fine-tune
embeddings \cite{cotterell2016morphological,vulic2017morph}
or to completely recompute them. The latter can be achieved using a model
trained to reproduce (or \emph{mimic}) the original
embeddings \cite{pinter2017mimicking}. However, these methods
only work if a word's meaning can at least partially be predicted from its form.

A closely related line of research is embedding learning for
\emph{novel words}, where the goal is to obtain embeddings
for previously unseen words from at most a handful of
observations. While most contemporary approaches exclusively
use context information for this task
\citep[e.g.][]{herbelot2017high,khodak2018carte},
\citet{schick2018learning} recently introduced the
\emph{form-context model} and showed that
joint learning from both surface
form and context
leads to
better performance.

The problem we address in this paper is that often, only
few of a word's contexts provide valuable information about its
meaning. Nonetheless, the current state of the art treats all contexts the
same.  We address this issue by introducing a more
intelligent mechanism of incorporating context into
mimicking: instead of using all contexts, we learn -- by way
of self-attention -- to pick a subset of especially informative and reliable
contexts. This mechanism is based on the observation that in many cases, reliable contexts for a given word tend to resemble each other. We call our proposed architecture \emph{attentive
  mimicking} (AM).
  
Our contributions are as follows:
(i) We introduce the attentive mimicking model. It produces
high-quality embeddings for rare and
medium-frequency words by attending to the most informative
contexts.
(ii) We propose a novel evaluation method based on
VecMap \cite{artetxe2018aaai} that allows us to easily evaluate
the embedding quality of low- and medium-frequency words.
(iii) We show that attentive mimicking improves word embeddings on various datasets.

\section{Related Work}

Methods to train surface-form models to mimic word
embeddings include those of
\citet{luong2013better} (morpheme-based) and
\citet{pinter2017mimicking} (character-level).
In the area of fine-tuning methods,
\citet{cotterell2016morphological} introduce a Gaussian
graphical model that incorporates morphological information into
word embeddings. \citet{vulic2017morph} retrofit embeddings
using a set of language-specific rules.  Models that
directly incorporate surface-form information into embedding
learning include fastText \cite{bojanowski2016enriching},
LexVec \cite{salle2018incorporating} and Charagram
\cite{wieting2016charagram}.

While many approaches to learning embeddings for novel words
exclusively make use of context information
\cite{lazaridou2017multimodal,herbelot2017high,khodak2018carte}, \citet{schick2018learning}'s form-context model combines
surface-form and context information.

\citet{ling2015not} also use attention in embedding
learning, but
their attention is \emph{within} a context (picking words),
not \emph{across} contexts (picking contexts).
Also, their attention is based only on word type and distance, not
on the more complex factors available in our attentive
mimicking model, e.g., the interaction with the word's surface form.


\section{Attentive Mimicking}

\subsection{Form-Context Model}
We briefly review
the architecture of the form-context model (FCM),
see \citet{schick2018learning} for more details.

FCM requires an embedding space of dimensionality $d$ that assigns high-quality embeddings $v \in \mathbb{R}^d$ to frequent words.
Given an infrequent or novel word $w$ and a set of contexts $\mathcal{C}$ in which it occurs, FCM can then be used to infer an embedding $v_{(w, \mathcal{C})}$ for $w$ that is appropriate for the given embedding space. This is achieved by first computing two distinct embeddings, one of which exclusively uses surface-form information and the other context information. The surface-form embedding, denoted $v_{(w,\mathcal{C})}^\text{form}$, is obtained from averaging over a set of $n$-gram embeddings learned by the model; the context embedding $v_{(w,\mathcal{C})}^\text{context}$ is obtained from averaging over all embeddings of context words in $\mathcal{C}$.

 The two embeddings are then
combined using a weighting coefficient $\alpha$ and a $d\times d$ matrix $A$, resulting in
the form-context embedding
\[
v_{(w,\mathcal{C})} = \alpha \cdot A v_{(w,\mathcal{C})}^\text{context} + (1 - \alpha) \cdot v_{(w,\mathcal{C})}^\text{form}\,.
\]
The weighing coefficient $\alpha$ is a function of both embeddings, modeled as
\[
\alpha = \sigma(u^\top [{v}_{({w}, \mathcal{C})}^\text{context};{v}_{({w}, \mathcal{C})}^\text{form}] + b)
\]
with $u\in \mathbb{R}^{2d}$, $b \in \mathbb{R}$ being
learnable parameters and $\sigma$ denoting the sigmoid function.

\subsection{Context Attention}

FCM pays equal attention to all contexts of a word but often,
only few contexts are actually suitable
for inferring the word's meaning. We introduce
\emph{attentive mimicking} (AM) to address this problem: we
allow 
our model to
assign different weights to contexts based on some measure
of their ``reliability''. To this end, let $\mathcal{C} =
\{C_1, \ldots, C_m\}$ where each $C_i$ is a multiset of
words.  We replace the context-embedding of FCM with a weighted embedding %
\[ %
v_{(w, \mathcal{C})}^\text{context} = %
\  \sum_{i=1}^m \rho(C_i, \mathcal{C})\cdot v_{C_i} %
\] %
where $v_{C_i}$ is  the average of the embeddings of words in $C_i$ and $\rho$ measures context reliability.

To obtain a meaningful measure of reliability, our key
observation is that reliable contexts typically agree with
many other contexts. Consider a word $w$ for which six out
of ten contexts contain words referring to sports. Due to
this high inter-context agreement, it is then reasonable to
assume that $w$ is from the same domain and,
consequently, that the four contexts not related to sports
are less informative. To
formalize this idea, we first define the
similarity between two contexts
as %
\[
s(C_1, C_2) =
\frac{(Mv_{C_1}) \cdot
(Mv_{C_2})^\top}{\sqrt{d}} %
\]
with $M \in \mathbb{R}^{d
  \times d}$ a learnable parameter, inspired by
\citet{Vaswani2017}'s
scaled dot-product attention.  We then define the reliability of a
context as
\[
\rho(C, \mathcal{C}) = \frac{1}{Z} \sum_{i=1}^m s(C, C_i)
\]
where $Z = \sum_{i=1}^m \sum_{j=1}^m s(C_i, C_j)$ is a normalization constant, ensuring that all weights sum to one.

The model is trained by randomly sampling words $w$ and contexts $\mathcal{C}$ from a large corpus and mimicking the original embedding of $w$, i.e., minimizing the squared distance between the original embedding and $v_{(w,\mathcal{C})}$.

\section{Experiments}

For our experiments, we follow the setup of
\citet{schick2018learning} and use the Westbury Wikipedia Corpus (WWC) \cite{shaoul2010westbury} for training of all embedding models. To obtain training instances $(w, \mathcal{C})$ for both FCM and AM, we sample words and contexts from the WWC based on their frequency, using only words that occur at least 100 times. We always train FCM and AM on skipgram embeddings \cite{Mikolov2013} obtained using Gensim \cite{rehurek_lrec}.

Our experimental setup differs from that of \citet{schick2018learning} in two respects: (i) Instead of using a fixed number of contexts for $\mathcal{C}$, we randomly sample between 1 and 64 contexts and (ii) we fix the number of training epochs to 5. The rationale behind our first modification is that we want our model to produce high-quality embeddings both when we only have a few contexts available and when there is a large number of contexts to pick from. We fix the number of epochs simply because our evaluation tasks come without development sets on which it may be optimized.

To evaluate our model, we apply a novel, intrinsic evaluation method that compares embedding spaces by transforming them into a common space (\secref{vecmap}). We also test our model on three word-level downstream tasks (\secref{sentiment}, \secref{nametyping}, \secref{chimeras}) to demonstrate its versatile applicability.

\subsection{VecMap}
\seclabel{vecmap}

We introduce a novel evaluation method that explicitly evaluates embeddings for rare and medium-frequency words by downsampling frequent words from the WWC to a fixed number of occurrences.\footnote{The VecMap dataset is publicly available at \url{https://github.com/timoschick/form-context-model}} We then compare ``gold'' skipgram embeddings obtained from the original corpus with embeddings learned by some model trained on the downsampled corpus. To this end, we transform the two embedding spaces into a common space using VecMap \cite{artetxe2018aaai}, where we provide all but the downsampled words as a mapping dictionary. Intuitively, the better a model is at inferring an embedding from few observations, the more similar its embeddings must be to the gold embeddings in this common space. We thus measure the quality of a model by computing the average cosine similarity between its embeddings and the gold embeddings.

As baselines, we train skipgram and fastText on the
downsampled corpus. We then train Mimick \cite{pinter2017mimicking} as well as both FCM and AM
on the skipgram embeddings. We also try a variant where the downsampled words are included in the training set (i.e., the mimicking models explicitly learn to reproduce their skipgram embeddings). This allows the model to learn representations of those words not completely from scratch, but to also make use of their original embeddings. Accordingly, we expect this variant to only be helpful if a word is not too rare, i.e. its original embedding is already of decent quality.
Table~\ref{results-vm} shows that for
words with a frequency below 32,  FCM and AM infer much better
embeddings than all baselines. The comparably poor performance of Mimick is consistent with the observation of \citet{pinter2017mimicking} that this method captures mostly syntactic information. Given four or more contexts,
AM leads to consistent improvements over FCM. The variants
that include downsampled words during training ($\dagger$) still outperform skipgram for 32 and more observations, but perform worse than the default models for less frequent words.

\begin{table}
  \centering
  {\small
  \setlength{\tabcolsep}{3pt}
  \newcolumntype{R}{>{\raggedleft\arraybackslash}X}
\begin{tabularx}{\linewidth}{lRRRRRRRR}
\toprule
& \multicolumn{8}{c}{\textbf{number of occurrences}} \\
\textbf{model} & 1 & 2 & 4 & 8 & 16 & 32 & 64 & 128 \\
\midrule 
skipgram  & 8.7 & 18.2 & 30.9 & 42.3 & 52.3 & 59.5 & 66.7 & \textbf{71.2} \\
fastText  & \textbf{45.4} & 44.3 & 45.7 & 50.0 & 55.9 & 56.7 & 62.6 & 67.7 \\
Mimick & 10.7 & 11.7 & 12.1 & 11.0 & 12.5 & 11.0 & 10.6 & 9.2 \\
FCM & 37.9 & \textbf{45.3} & 49.1 & 53.4 & \textbf{58.3} & 55.4 & 59.9 & 58.8 \\
AM & 38.0 & 45.1 & \textbf{49.6} & \textbf{53.7} & \textbf{58.3} & 55.6 & 60.2 & 58.9 \\
FCM$^\dagger$ & 32.3 & 36.9 & 41.9 & 49.1 & 57.4 & 59.9 & 67.3 & 70.1 \\
AM$^\dagger$ & 32.8 & 37.8 & 42.8 & 49.8 & 57.7 & \textbf{60.5} & \textbf{67.6} & 70.4 \\
\bottomrule
\end{tabularx}}
\caption{Average cosine similarities for the VecMap evaluation, scaled by a factor of 100. $\dagger$: Downsampled words were included in the training set.
\label{results-vm}}
\end{table}

\begin{table}
  \centering
  {\small
  \newcolumntype{R}{>{\raggedleft\arraybackslash}X}
\begin{tabularx}{\linewidth}{lRRRRR}
\toprule
& \multicolumn{5}{c}{\textbf{maximum word frequency}} \\
\textbf{model} & \multicolumn{1}{c}{10} & \multicolumn{1}{c}{50} & \multicolumn{1}{c}{100} & \multicolumn{1}{c}{500} & \multicolumn{1}{c}{1000}  \\
\midrule 
skipgram & \textminus0.16 & $0.21$ & $0.33$ & $0.55$ & $\mathbf{0.66}$ \\
fastText & \textminus0.20 & $0.10$ & $0.23$ & $0.50$ & $0.61$ \\
Mimick & 0.00 & 0.01 & \textminus0.03 & 0.40 & 0.56 \\
FCM & $0.21$ & $0.37$ & $0.37$ & $0.55$ & $0.63$ \\
AM & $\mathbf{0.27}$ & $\mathbf{0.39}$ & $\mathbf{0.40}$ & $\mathbf{0.56}$ & $0.64$ \\
\bottomrule
\end{tabularx}}
\caption{Spearman's $\rho$ for various approaches on \mbox{SemEval2015} Task 10E 
\label{results-se}}
\end{table}

\begin{table*}[t]
  \centering
  {\small
  \newcolumntype{R}{>{\raggedleft\arraybackslash}X}
\begin{tabularx}{\linewidth}{lRRRRRRRRRRRRRR}
\toprule
 & \multicolumn{2}{c}{$f{\,=\,}1$} & \multicolumn{2}{c}{$f{\,\in\,}[2,4)$} & \multicolumn{2}{c}{$f{\,\in\,}[4,8)$} & \multicolumn{2}{c}{$f{\,\in\,}[8,16)$} & \multicolumn{2}{c}{$f{\,\in\,}[16,32)$} & \multicolumn{2}{c}{$f{\,\in\,}[32,64)$} 
& \multicolumn{2}{c}{$f{\,\in\,}[1,100]$} \\
\cmidrule(lr){2-3}\cmidrule(lr){4-5}\cmidrule(lr){6-7}\cmidrule(lr){8-9}\cmidrule(lr){10-11}\cmidrule(lr){12-13}\cmidrule(lr){14-15}
\textbf{model} & acc & F1& acc & F1& acc & F1& acc & F1& acc & F1& acc & F1& acc & F1 \\
\midrule 
skipgram & 0.0 & 2.6 & 2.2 & 7.8 & 11.5 & 30.7 & 44.7 & 64.5 & 37.8& 59.4 & \textbf{35.0} & \textbf{59.7} & 33.5 & 58.3 \\
fastText & 44.6 & 51.1 & 50.5 & 65.1 & 48.4 & 62.9 & 44.3 & 59.6 & 34.1 & 53.5 & 29.8 & 55.7 & 31.4 & 56.4 \\
Mimick & 0.0 & 0.0 & 0.0 & 0.0 & 0.0 & 0.0 & 1.0 & 4.0 & 1.0 & 1.0 & 3.9 & 14.4 & 4.2 & 14.8 \\
FCM & 86.5 & 88.9 & 76.9 & 85.1 & \textbf{72.0} & \textbf{81.8} & 57.7 & 68.5 & 36.0 & 54.2 & 27.7 & 52.5 & 30.1 & 53.4 \\
AM & \textbf{87.8} & \textbf{90.7} & \textbf{79.1} & \textbf{86.5} & \textbf{72.0} & 80.9 & 59.5 & \textbf{70.9} & 37.8 & 56.1 & 28.9 & 53.4 & 31.1 & 54.5 \\
AM+skip & \textbf{87.8} & \textbf{90.7} & \textbf{79.1} & \textbf{86.5} & \textbf{72.0} & 81.6 & \textbf{60.1} & \textbf{70.9} & \textbf{40.7} & \textbf{59.9} & \textbf{35.0} & \textbf{59.7} & \textbf{36.8} & \textbf{60.5} \\
\bottomrule
\end{tabularx}}
\caption{Results on the Name Typing dataset for various word frequencies $f$. The model that uses a linear combination of AM embeddings with skipgram is denoted AM+skip.
\label{results-nt-mlp}}
\end{table*}

\subsection{Sentiment Dictionary}
\seclabel{sentiment}
We follow the experimental setup of \citet{rothe2016ultradense} and fuse Opinion lexicon \cite{hu2004mining} and the NRC Emotion lexicons \cite{mohammad2013crowdsourcing} to obtain a training set of words with binary sentiment labels. On that data, we train a logistic regression model to classify words based on their embeddings. For our evaluation, we then use SemEval2015 Task 10E where words are assigned a sentiment rating between 0 (completely negative) and 1 (completely positive) and use Spearman's $\rho$ as a measure of similarity between 
gold and predicted ratings.

We train logistic regression models on both skipgram and
fastText embeddings and, for testing, replace skipgram embeddings by embeddings inferred from the mimicking models.  
Table~\ref{results-se} shows  that for rare
and medium-frequency words, AM again
outperforms all other models.

\subsection{Name Typing}
\seclabel{nametyping}
We use
\citet{yaghoobzadeh2018evaluating}'s 
name typing dataset for the task of predicting
the fine-grained named entity types of a word, e.g., \textsc{president}
and \textsc{location} for ``Washington''.
We train a logistic regression model using the same setup as
in \secref{sentiment}
and evaluate on all words from the test set that occur
$\leq$100 times in WWC.
Based on results in \secref{vecmap},
where AM only improved representations for words occurring
fewer than 32 times, we also try the variant \emph{AM+skip}
that, in testing, replaces $v_{(w,\mathcal{C})}$ with   the linear combination
\[
\hat{v}_w = \beta(f_w) \cdot v_{(w,\mathcal{C})} + (1 - \beta(f_w)) \cdot v_w
\]
where $v_w$ is the skipgram embedding of $w$, $f_w$ is the frequency of $w$ and $\beta(f_w)$ scales linearly from $1$ for $f_w = 0$ to $0$ for $f_w = 32$.

Table~\ref{results-nt-mlp}
gives accuracy and micro F1 for several word frequency
ranges. In accordance with results from previous
experiments,  AM performs drastically better than the baselines for up to 16 occurrences. Notably, the linear combination of skipgram and  AM achieves by far the best overall results.

\subsection{Chimeras}
\seclabel{chimeras}
The Chimeras (CHIMERA) dataset \cite{lazaridou2017multimodal} consists
of similarity scores for pairs of made-up words and regular
words.
CHIMERA
provides only six contexts for each
made-up word, so 
it is not ideal for evaluating
our model. Nonetheless, 
we can still use it to analyze the
difference of FCM (no attention) and AM (using attention).
As the surface-form of the made-up words was constructed randomly and thus carries no meaning at all, we restrict ourselves to
the context parts of FCM and AM (referred to as FCM-ctx and
AM-ctx).
We use the test set of
\citet{herbelot2017high} and compare the given similarity scores with the cosine similarities of the corresponding word embeddings, using FCM-ctx and AM-ctx to obtain embeddings for the made-up words.
Table~\ref{results-ch} gives
Spearman's $\rho$ for our model and various baselines; baseline results are adopted from \citet{khodak2018carte}. We do not report results for Mimick as its representations for novel words are entirely based on their surface form.
While
AM performs worse than previous methods for 
2--4 sentences, it drastically improves
over the best result currently published for 6
sentences. 
Again, context attention
consistently improves results: AM-ctx performs better than FCM-ctx, regardless of the number of contexts.
Since A La Carte \cite{khodak2018carte}, the method performing best for 2--4 contexts, is
conceptually similar to FCM,
it most likely would similarly benefit from context attention.

\begin{table}
  \centering
  {\small
  \newcolumntype{R}{>{\raggedleft\arraybackslash}X}
\begin{tabularx}{\linewidth}{lRRR}
\toprule
\textbf{model} & 2 sent. & 4 sent. & 6 sent. \\
\midrule 
skipgram & 0.146 & 0.246 & 0.250 \\
additive & \textbf{0.363} & 0.370 & 0.360 \\
additive \textminus\,sw & 0.338 & 0.362 & 0.408 \\
Nonce2Vec & 0.332 & 0.367 & 0.389 \\
A La Carte & \textbf{0.363} & \textbf{0.384} & 0.394 \\
FCM-ctx & 0.337 & 0.359 & 0.422 \\
AM-ctx  & 0.342 & 0.376 & \textbf{0.436} \\
\bottomrule
\end{tabularx}}
\caption{Spearman's $\rho$ for the Chimeras task given 2, 4 and 6 context sentences for the made-up word
\label{results-ch}}
\end{table}

\begin{table}
  \centering
  {\small
    \setlength{\tabcolsep}{3pt}
\begin{tabularx}{\linewidth}{rXr}
\toprule
\multicolumn{2}{l}{sentence} & \multicolumn{1}{c}{$\rho$}\\
\midrule 
$\bullet$ & i doubt if we ll ever hear a man play a \textbf{petfel} like that again & 0.19 \\
$\bullet$ & also there were some other assorted instruments including a \textbf{petfel} and some wind chimes & 0.31 \\
$\bullet$ & they finished with new moon city a song about a suburb of drem which featured beautifully controlled \textbf{petfel} playing from callum & 0.23 \\
$\bullet$ & a programme of jazz and classical music showing the \textbf{petfel} as an instrument of both musical genres & 0.27 \\
\bottomrule
\end{tabularx}}
\caption{Context sentences and corresponding attention weights for the made-up word ``petfel''
\label{sentences-ch}}
\end{table}

While the effect of context attention is more pronounced
when there are many contexts available, we still perform a
quantitative analysis of one exemplary instance of
CHIMERA to better understand what AM learns; we consider the made-up word ``petfel'', a
combination of ``saxophone'' and ``harmonica'', whose occurrences are shown in Table~\ref{sentences-ch}. The model attends
most to sentences (2) and (4); consistently, the embeddings
obtained from those sentences are very similar. Furthermore, of
all four sentences, these two are the ones best
suited for a simple averaging model as they contain informative, frequent words
like ``instrument'', ``chimes'' and ``music''.

\section{Conclusion}

We have introduced attentive mimicking (AM) and showed that attending to informative and reliable contexts
improves representations of rare and medium-frequency words
for a diverse set of evaluations.

In future work, one might investigate whether attention
mechanisms on the word level
\cite[cf.][]{ling2015not}
can further improve
the model's performance. Furthermore, it would be interesting to
investigate whether the proposed architecture is also
beneficial for languages typologically different from
English, e.g., morphologically rich languages.

\section*{Acknowledgments}
This work was funded by the European Research Council (ERC \#740516).
We would like to thank the anonymous reviewers
for their helpful comments.

\bibliography{literatur}
\bibliographystyle{acl_natbib}

\clearpage

\appendix
\section{Experimental Details}

In all of our experiments, we train embeddings on the Westbury Wikipedia Corpus (WWC) \cite{shaoul2010westbury}. For skipgram, we use Gensim \cite{rehurek_lrec} and its default settings with two exceptions:
\begin{itemize}
\item We leave the minimum word count at 50, but we explicitly include all words that occur in the test set of our evaluation tasks, even if they occur less than 50 times in the WWC.
\item We increase the dimensionality $d$ of the embedding space; the values of $d$ chosen for each experiment are mentioned below.
\end{itemize}

For experiments in which we use fastText, we use the default parameters of the implementation by \citet{bojanowski2016enriching}. To evaluate the Mimick model by \citet{pinter2017mimicking}, we use their implementation and keep the default settings.

To obtain training instances for the attentive mimicking model, we use the same setup as \citet{schick2018learning}: we use only words occurring at least 100 times in the WWC and if a word $w$ has a total of $f(w)$ occurrences, we train on it $n(w)$ times for each epoch, where
\[
n(w) = \min( \lfloor \frac{f(w)}{100} \rfloor, 5)\,.
\]
 We restrict each context of a word to at most 25 words on its left and right, respectively. While \citet{schick2018learning} use a fixed number of 20 contexts per word during training, we instead randomly sample between 1 and 64 contexts. We do so for both the form-context model and the attentive mimicking model as we found this modification to generally improve results for both models. For all experiments, we train both the form-context model and the attentive mimicking model for 5 epochs using the Adam optimizer \cite{kingma2014adam} with an initial learning rate of 0.01 and a batch size of 64.

\subsection*{VecMap}

The test set for the VecMap evaluation was created using the following steps:
\begin{enumerate}
\item We sample 1000 words from the lowercased and tokenized WWC that occur at least 1000 times therein, contain only alphabetic characters and at least two characters.
\item We evenly distribute the 1000 words into 8 buckets $B_0, \ldots, B_7$ such that each bucket contains 125 words.
\item We downsample each word $w$ in bucket $B_i$ to exactly $2^i$ randomly chosen occurrences.
\end{enumerate}
For the variants of AM and FCM where the downsampled words are included in the training set, in every epoch we construct 5 training pairs  $(w, \mathcal{C}_1), \ldots, (w, \mathcal{C}_5)$  for each downsampled word $w$.
For training of both skipgram and fastText, we use 400-dimensional embeddings. 

\subsection*{Sentiment Dictionary}

To obtain the training set for the Sentiment Dictionary evaluation, we fuse Opinion lexicon \cite{hu2004mining} and the NRC Emotion lexicons \cite{mohammad2013crowdsourcing} and remove all words that occur less than 100 times in the WWC corpus. From the SemEval2015 Task 10E data set, we remove all non-alphanumeric characters and all words that have less than 2 letters. We do so as the test set contains many hashtags, giving an unfair disadvantage to our baseline skipgram model as it makes no use of surface-form information.

We use 300-dimensional embeddings and train the logistic regression model for 5 epochs using the Adam optimizer \cite{kingma2014adam} with an initial learning rate of $0.01$.

\subsection*{Name Typing}

We use the same setup as for the Sentiment Dictionary experiment. That is, we use 300-dimensional embeddings and train the logistic regression model for 5 epochs using the Adam optimizer \cite{kingma2014adam} with an initial learning rate of $0.01$.

\subsection*{Chimeras}

Following \citet{herbelot2017high}, we use 400-dimensional embeddings for the Chimeras task.

\begin{table*}
  \centering
  {\small
  \setlength{\tabcolsep}{6pt}
  \newcolumntype{R}{>{\raggedleft\arraybackslash}X}
\begin{tabularx}{\linewidth}{llllllll}
\toprule
\textbf{model} & skipgram & fastText & Mimick & FCM & AM & FCM$^\dagger$ & AM$^\dagger$ \\
\midrule 
skipgram & -- & 64,128 & 2,4,8,16,32,64,128 & 32,64,128 & 32,64,128 & -- & --  \\
fastText & 1,2,4,8,16 & -- & 1,2,4,8,16,32,64,128 & 1,128 & 1,128 & 1,2,4 & 1,2,4  \\
Mimick & -- & -- & -- & -- & -- & -- & --  \\
FCM & 1,2,4,8,16 & 8 & 1,2,4,8,16,32,64,128 & -- & -- & 1,2,4,8 & 1,2,4,8  \\
AM & 1,2,4,8,16 & 8 & 1,2,4,8,16,32,64,128 & 1,4 & -- & 1,2,4,8,16 & 1,2,4,8,16  \\
FCM$^\dagger$ & 1,2,4,8,16 & 32,64,128 & 1,2,4,8,16,32,64,128 & 32,64,128 & 32,64,128 & -- & --  \\
AM$^\dagger$ & 1,2,4,8,16,64 & 32,64,128 & 1,2,4,8,16,32,64,128 & 32,64,128 & 32,64,128 & 2,4,8,32,64,128 & --  \\
\bottomrule
\end{tabularx}}
\caption{Significance results for the VecMap evaluation. Each cell lists the numbers of word occurrences for which the model of the row performs significantly better than the model of the column ($p < 0.05$). For example, FCM is significantly better than skipgram for 1, 2, 4, 8 and 16 contexts.
\label{results-vm-significance}}
\end{table*}

\begin{table*}
  \centering
  {\small
    \setlength{\tabcolsep}{8pt}
  \newcolumntype{R}{>{\raggedleft\arraybackslash}X}
\begin{tabularx}{\linewidth}{lllllll}
\toprule
\textbf{model} & skipgram & fastText & Mimick & FCM & AM & AM+skip \\
\midrule 
skipgram & -- & $f_4$,$f_5$,$f_6$ & $f_2$,$f_3$,$f_4$,$f_5$,$f_6$ & $f_5$,$f_6$ & $f_5$,$f_6$ & -- \\
fastText & $f_0$,$f_1$,$f_2$ & -- & $f_0$,$f_1$,$f_2$,$f_3$,$f_4$,$f_5$,$f_6$ & $f_5$,$f_6$ & -- & -- \\
Mimick & -- & -- & -- & -- & -- & --  \\
FCM & $f_0$,$f_1$,$f_2$,$f_3$ & $f_0$,$f_1$,$f_2$,$f_3$ & $f_0$,$f_1$,$f_2$,$f_3$,$f_4$,$f_5$,$f_6$ & -- & -- & --  \\
AM & $f_0$,$f_1$,$f_2$,$f_3$ & $f_0$,$f_1$,$f_2$,$f_3$,$f_4$ & $f_0$,$f_1$,$f_2$,$f_3$,$f_4$,$f_5$,$f_6$ & $f_4$,$f_5$,$f_6$ & -- & -- \\
AM+skip & $f_0$,$f_1$,$f_2$,$f_3$,$f_4$,$f_6$ & $f_0$,$f_1$,$f_2$,$f_3$,$f_4$,$f_5$,$f_6$ & $f_0$,$f_1$,$f_2$,$f_3$,$f_4$,$f_5$,$f_6$ & $f_4$,$f_5$,$f_6$ & $f_4$,$f_5$,$f_6$ & -- \\

\bottomrule
\end{tabularx}}
\caption{Significance results for the Name Typing task. Each cell lists the frequency intervals for which the model of the row performs significantly better than the model of the column ($p < 0.05$) with regards to micro accuracy. We use abbreviations $f_i = [2^i, 2^{i+1})$ for $0 \leq i \leq 5$ and $f_6 = [1,100]$. 
\label{results-nt-significance}}
\end{table*}

\section{Significance Tests}

We perform significance tests for the results obtained on both the VecMap and the Name Typing dataset. 

For VecMap, given two models $m_1$ and $m_2$, we count the number of times that the embedding assigned to a word $w$ by $m_1$ is closer to the gold embedding of $w$ than the embedding assigned by $m_2$; we do so for each number of occurrences separately. Based on the so-obtained counts, we perform a binomial test whose results are shown in Table~\ref{results-vm-significance}. As can be seen, both FCM and AM perform significantly better than the original skipgram embeddings for up to 16 contexts, but the difference between FCM and AM is only significant given one or four contexts. However, for the variants
that include downsampled words during training, AM$^\dagger$ (using attention) is significantly better than FCM$^\dagger$ (without attention) given more than one context.

For the Name Typing dataset, we compare models based on their micro accuracy, ignoring all dataset entries for which both models perform equally well. Again, we consider all frequency ranges separately. Results of the binomial test for significance can be seen in Table~\ref{results-nt-significance}. The best-performing method, AM+skip, is significantly better than skipgram, fastText and Mimick for almost all frequency ranges. AM is significantly better than FCM only when there is a sufficient number of contexts.

\end{document}